\documentclass{ieeeaccess}
\usepackage{amsmath,amssymb,amsfonts}
\usepackage{algorithmic}
\usepackage{graphicx}
\usepackage{textcomp}
\usepackage{booktabs}
\usepackage{float}
\usepackage{url}
\usepackage{blindtext}
\usepackage[style=ieee,defernumbers=true,mincitenames=1,maxcitenames=2,backend=biber]{biblatex}
\usepackage{hyperref}
\usepackage{caption}
\usepackage{subcaption}
\usepackage[table,xcdraw]{xcolor}

\begin{document}

\title{Data-Driven Discovery of Emergent Dynamics in Reaction–Diffusion Systems from Sparse and Noisy Observations}

\author{\uppercase{Saumitra Dwivedi}\authorrefmark{1,2},
\uppercase{Ricardo da Silva Torres\authorrefmark{3}},
\uppercase{Ibrahim A. Hameed}\authorrefmark{4},
\uppercase{Gunnar Tufte}\authorrefmark{2},
\uppercase{Anniken Susanne T. Karlsen}\authorrefmark{1}}
\address[1]{Department of ICT and Natural Sciences, Norwegian University of Science and Technology, Ålesund, Norway}
\address[2]{Department of Computer Science, Norwegian University of Science and Technology, Trondheim, Norway}
\address[3]{Artificial Intelligence Group, Wageningen University and Research, Wageningen, The Netherlands}
\address[4]{Department of Mechanical Engineering and Technology Management, Norwegian University of Life Sciences, Ås, Norway}

\corresp{Corresponding author: Saumitra Dwivedi (e-mail: saumitra.dwivedi@ntnu.no).}

\begin{abstract}
Data-driven discovery of emergent dynamics is gaining popularity, particularly in the context of reaction-diffusion systems. These systems are widely studied across various fields, including neuroscience, ecology, epidemiology, and several other subject areas that deal with emergent dynamics. A current challenge in the discovery process relates to system identification when there is no prior knowledge of the underlying physics. We attempt to address this challenge by learning Soft Artificial Life (Soft ALife) models, such as Agent-based and Cellular Automata (CA) models, from observed data for reaction-diffusion systems. 
In this paper, we present findings on the applicability of a conceptual framework, the Data-driven Rulesets for Soft Artificial Life (DRSALife) model, to learn Soft ALife rulesets that accurately represent emergent dynamics in a reaction-diffusion system from observed data. This model has demonstrated promising results for Elementary CA Rule 30, Game of Life, and Vicsek Flocking problems in recent work. To our knowledge, this is one of the few studies that explore machine-based Soft ALife ruleset learning and system identification for reaction-diffusion dynamics without any prior knowledge of the underlying physics. 
Moreover, we provide comprehensive findings from experiments investigating the potential effects of using noisy and sparse observed datasets on learning emergent dynamics. Additionally, we successfully identify the structure and parameters of the underlying partial differential equations (PDEs) representing these dynamics. Experimental results demonstrate that the learned models are able to predict the emergent dynamics with good accuracy (74\%) and exhibit quite robust performance when subjected to Gaussian noise and temporal sparsity. The methodology presented in this paper offers valuable approaches for identifying and characterizing system dynamics across various domains, including ecology, epidemiology, biology, and others.
\end{abstract}

\begin{keywords}
Reaction-Diffusion Systems, Soft Artificial Life, System Identification, Machine Learning, Complex Systems, Cellular Automata, Emergent Dynamics, Data-driven Modeling, Agent-based Models, Ruleset Learning
\end{keywords}

\maketitle
\section{Introduction}
\label{sec:intro}
Emergent spatiotemporal dynamics are observed quite often in nature. An example of that can be seen in reaction-diffusion (RD) dynamics occurring and studied in neuroscience \cite{Banerjee2022,Crevat2019,Denizot2022}, chemistry \cite{WANG2024938,Soh2010,Mann2009}, ecology \cite{Chang2023,Belik2011,ChrisCosner2014}, epidemiology \cite{Wang2021,WANG20122240}, and more. Studies like \cite{ABBAS2025386,Hou2025,Sun2013} show the use of RD models to predict and control growth patterns in vegetation. Such use of RD models is relevant to understanding plant growth behavior, their behavior regarding competing for resources, and their reaction to environmental factors in an ecosystem. In addition, an example of the use of the RD process is in chemistry \cite{WANG2024938,Mann2009} where the RD process is used to achieve spatiotemporal control on the outcome of the overall chemical processes, thereby achieving the preferred state of matter.

Clearly understanding the nature of such spatiotemporal dynamics is crucial. The way to gain such understanding is often maneuvered around the identification of governing partial differential equations (PDEs) representing/modeling the dynamics. A promising research direction recently relies on employing data-driven discovery by system identification of emergent dynamics in an RD system.

Recent methods related to machine learning have led to a boom in data-driven discovery \cite{sindy2016,Rudy2017}, wherein data-driven methods are used to learn spatiotemporal dynamics from observed data, to provide opportunities for faster computation \cite{Kochkov2021,wang2020physicsinformeddeeplearningturbulent}, system identification and equation discovery \cite{sindy2016,Rudy2017}, physics-informed learning \cite{Rao2023,wang2020physicsinformeddeeplearningturbulent}, and more. Authors such as \textcite{Rao2023,Noordijk2024} recognize mechanistic modeling (finding underlying PDEs) as a major challenge in relation to modeling complex spatiotemporal dynamics, and argue that learning dynamics from observed data using data-driven methods is greatly helped by prior information/knowledge on the structure of physics behind the dynamics. However, in many applications, prior information about the structure of physical dynamics might not be available to guide the discovery process. Learning emergent dynamics with an unknown prior understanding of underlying physics still remains a challenge. We investigate an approach to learn emergent dynamics using sparse and noisy observed data for an RD spatiotemporal dynamical system, while also identifying the structure and parameters of the underlying PDEs.

The findings presented in this paper are based on a study focused on exploring system identification pertaining to an RD system and investigating the applicability of a recently formulated data-driven conceptual model~\cite{Dwivedi2025}, on an RD system. This conceptual model is referred to as the Data-driven Rulesets for Soft Artificial Life (DRSALife) model in what follows. The DRSALife model was formulated to help construct data-driven Soft Artificial Life (Soft ALife) methods like Agent-Based Models (ABMs) and Cellular Automata (CA), to model and simulate emergent spatiotemporal dynamics. To the best of our knowledge, this is one of the few studies focused on investigating Soft ALife machine-based ruleset learning and system identification for RD dynamics with unknown prior information on underlying physics.

Our study addresses the research question: \textbf{What are the implications of using the DRSALife model for learning the underlying dynamics of a reaction-diffusion system and identifying the parameters of a representative partial differential equation?} The investigation for the research question is structured into experiments wherein we investigate the robustness of using the DRSALife model towards noise, sparsity, and data observability, which are common issues usually found in real datasets pertaining to RD systems.

\section{The Reaction Diffusion (RD) system}
\label{sec:background-concepts}

\begin{figure}[]
    \centering
    \includegraphics[width=\linewidth]{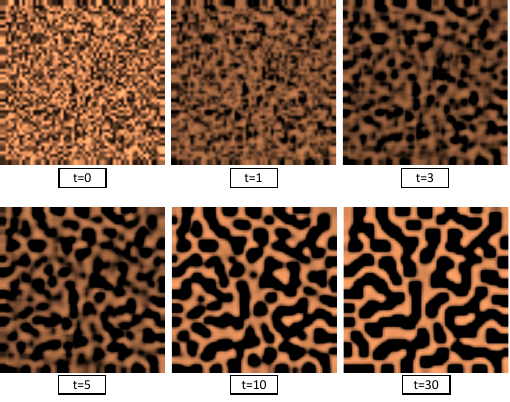}
    \caption{This figure shows an example of the formation of Turing patterns with time (six-timestep snapshots shown in this figure) as observed in chemical and biological systems. The parameters used to simulate these snapshots can be found in Section \ref{sec:Dataset_decription}.}
    \label{fig:turingPatterns}
\end{figure}

\begin{figure*}[!t]
    \centering
    \includegraphics[width=\linewidth]{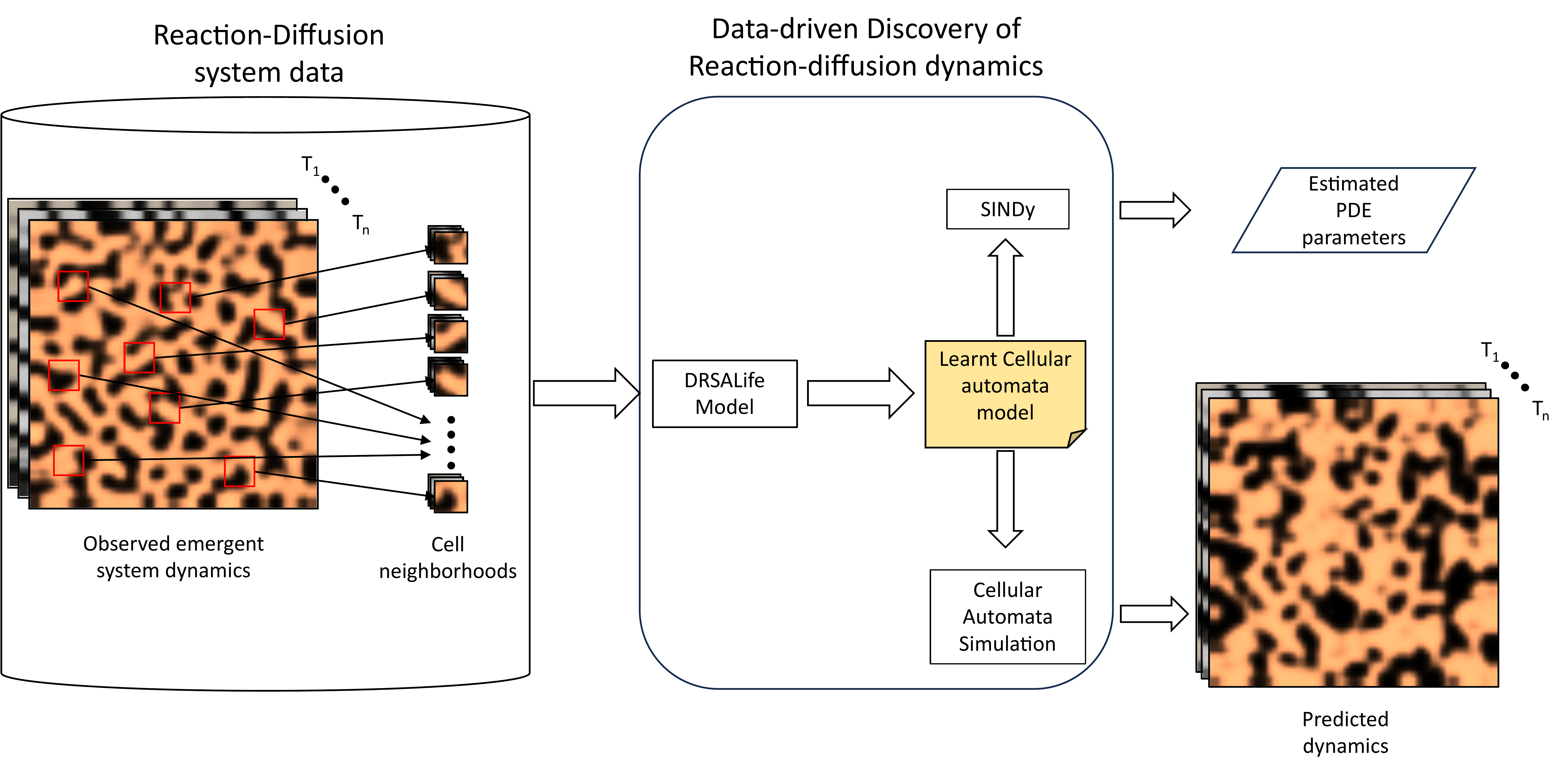}
    \caption{The figure depicts the overall process of data-driven discovery of reaction diffusion dynamics, described in the paper. Here, the reaction-diffusion dynamics observed at cell neighborhoods is used as input to the DRSALife model, to learn the emergent behavior at neighborhood level (or low level), thereby producing a data-driven Cellular Automata model. The learned Cellular Automata model is then used to predict dynamics on system level (or high level), and estimate parameters of a representative PDE by using the learned model along with Sparse Identification of Nonlinear Dynamical Systems (SINDy)~\cite{sindy2016}.}
    \label{fig:CA_pipeline}
\end{figure*}

\begin{figure*}[!t]
    \centering
    \includegraphics[width=0.8\linewidth]{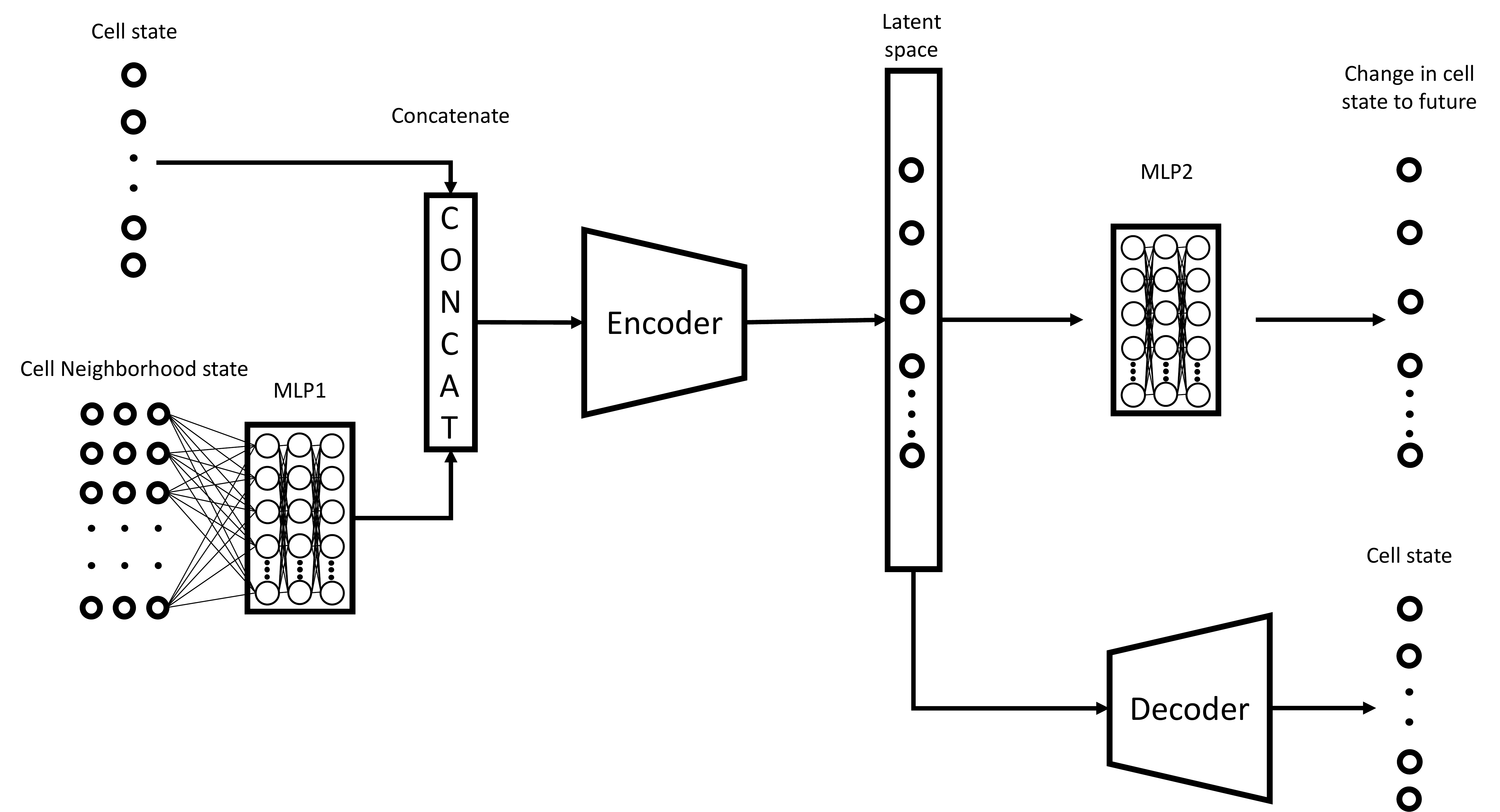}
    \caption{Proposed Artificial Neural Network pipeline used to obtain Feature expansion and consequently one-step prediction for CA model. The pipeline takes the state of a cell and its neighborhood as input and outputs the change in state of the cell.}
    \label{fig:ANN}
\end{figure*}

This section briefly introduces the underlying concepts used and discussed in this paper.

In the field of mathematical biology, the FitzHugh-Nagumo (FHN) RD model proposed an approach to represent the dynamics in nerve impulse propagation and chemical reactions \cite{FitzHugh1955,FITZHUGH1961445,Nagumo1962}. The FHN model was independently developed by \textcite{FitzHugh1955,FITZHUGH1961445} and \textcite{Nagumo1962}, wherein the model represents the Hodgkin-Huxley equations~\cite{Hodgkin_Huxley1952} in a two-dimensional (2D) dynamical system. The two dimensions or variables are the activator and inhibitor, representing the features of a neuron's action potential mechanism.

The central dynamics of the FHN model can be outlined by the following equations \cite{Winfree1991}:

\begin{equation}
    \frac{du}{dt} = \rho(u) - v + D_u\Delta u
\end{equation}

\begin{equation}
    \frac{dv}{dt} = \epsilon(u - \gamma v) + D_v\Delta v
\end{equation}

\noindent where, $u$ represents the activator, while $v$ is the inhibitor. The non-linear function $\rho(u)$ typically has a cubic shape, which allows for the characteristic excitable dynamics. For instance, $\rho(u) = u-u^3/3$ in \cite{FITZHUGH1961445}. The parameters $D_u$ and $D_v$ are diffusion coefficients, and $\epsilon$ and $\gamma$ are constants denoting the timescale separation and coupling strength between $u$ and $v$, respectively.

In 1952, \textcite{Turingpatterns1952} proposed that under certain conditions, a stable, homogeneous state can be destabilized through diffusion-driven instability, leading to the spontaneous formation of spatial patterns. Turing's work led to the concept of Turing patterns (an example is shown in Figure \ref{fig:turingPatterns}), much before they were observed in chemical and biological systems by Fitzhugh and Nagumo. These patterns can manifest as spots, stripes, or labyrinthine designs and are thought to be behind many natural patterns such as animal coat markings, fish skin pigmentation, and even aspects of tissue morphogenesis~\cite{RAMOS20243165,Turing70years}.

Recent works by \textcite{ABBAS2025386,Hou2025,Chang2023,Woolley2021} show that understanding Turing patterns through models like FHN has profound implications. In materials science, insights from the Turing patterns guide the design of self-organizing materials \cite{Xiang2022,Chehami2023}. Furthermore, insights gained by using the Turing patterns-based modeling in excitable media have implications for the development of cardiac tissue engineering and the control of excitable tissue behavior, thus bearing potential biomedical applications~\cite{Fatoyinbo2022,Tsyganov2014}.

\section{Materials and Methods}
In this section, we detail the overall method and experimental approach used in our study.

\subsection{Data-driven discovery of Reaction-Diffusion dynamics}
\label{sec:method}
Figure \ref{fig:CA_pipeline} depicts the overall process of our proposed approach, wherein the RD dynamics seen in cell neighborhoods serve as input for the DRSALife model (described in section \ref{sec:DRSALife}), allowing it to learn the emergent behaviors at the neighborhood (or low) level. This results in the formation of a data-driven CA model. Subsequently, the learned CA model is utilized to forecast dynamics at the system (or high) level and to estimate parameters of a representative PDE by using the learned model with SINDy (described in section \ref{sec:SINDy}). The two main component methods used in our approach, i.e., DRSALife model and SINDy, are described in Sections \ref{sec:DRSALife} and \ref{sec:SINDy}, respectively.

\subsubsection{Data-driven Rulesets for Soft Artificial Life (DRSALife) model}
\label{sec:DRSALife}
The DRSALife model aims to help construct Data-driven Soft Artificial Life methods like Agent-based models (ABM) and Cellular Automata (CA), to model and simulate emergent spatiotemporal dynamics using observed data. The DRSALife model was inspired by \textcite{Gershenson2023}, according to whom emergent behavior happens at different levels/scales of the system, i.e., gradual or slow emergent behavior occurs at a high level/scale of the system, whereas fast and non-linear behavior takes place at a low level/scale of the system. This model conceptualizes an approach to learn emergent behavior from observed data at the different levels/scales of the system.

In our study, we apply the DRSALife model in the context of an RD system. Firstly, we map the RD system dynamics to a CA model (described in the next paragraphs). Then we use Artificial Neural Network (ANN)-based feature transformation (also described in the next paragraphs) to model the emergent dynamics observed at low/fast levels (cell neighborhoods). This results in a data-driven CA model representative of the emergent RD dynamics.

\paragraph{Cellular Automata}
\label{sec:CA}
We consider a cellular automaton (CA) consisting of $N$ cells, each cell representing a part of the system. The cells evolve over discrete time steps through local interactions, modeling the dynamics of an RD system. Let $X_i(t)$ denote the state of cell $i$ at time $t$. For a two-species system, the state is given by the concentrations of two chemical species: $X_i(t) = (u_i(t), v_i(t))$ where $u$ represents the concentration of the first chemical species (e.g., an activator), and $v$ represents the concentration of the second chemical species (e.g., an inhibitor).

The full system state at time $t$ is constructed by stacking the individual cell states:
\[
X(t) = [X_1(t), X_2(t), \dots, X_N(t)].
\]

Each cell interacts with its four immediate neighbors (up, down, left, right), forming a five-point stencil. Let ${X_p}_i(t)$ denote the neighborhood state for cell $i$, such that ${X_p}_i(t) = [{X_{up}}_i(t),{X_{bottom}}_i(t),{X_{left}}_i(t),{X_{right}}_i(t)]$  and let:
\[
X_p(t) = [{X_p}_1(t), {X_p}_2(t), \dots, {X_p}_N(t)]
\]
represent the stacked neighborhood states for all cells.

We assume that each cell's update depends only on its own state and that of its neighbors. The update rule for the discrete-time system is given by:
\[
X(t+1) = X(t) + f(X(t), X_p(t)),
\]
where $f(\cdot)$ defines the transition dynamics of the CA and $t$ represents a timestep of the discrete-time system.

\paragraph{Artificial Neural Network based Feature Transformation}
We consider a Multi-Layer Perceptron (MLP)-based feature space expansion$/$ higher order transformation wherein we exploit an expanded feature space to predict a one-step change in the state of the system, i.e., we approximate $f$ in $\dot{X(t)} = f(X(t), X_p(t))$. Correspondingly, the ANN takes $(X, X_p)$ as input and outputs $\dot{X}$.

Figure~\ref{fig:ANN} depicts the developed ANN, comprising the two main network components of Reconstruction and Prediction. All proposed ANN components were implemented using MLP. 

\begin{enumerate}
    \item Reconstruction: To obtain a higher-order subspace, we implement a network architecture similar to an autoencoder. Usually, autoencoders are used to obtain a compressed latent space. In our implementation, the latent space size is higher than the original state space. The encoder maps the concatenated space of cell and cell neighborhood $(X, X_k)$ to a higher-dimensional latent space, and the decoder maps the latent space back to cell space $(X)$.
    \item Prediction: We consider the prediction of state change in a cell, given a neighborhood setting. We model this by mapping the high-dimensional latent space through a nonlinear function (using an MLP).
\end{enumerate}

Table~\ref{tab:network_phi} provides details of the architecture of individual network components, MLP1, Encoder, Decoder, and MLP2.

\begin{table}[ht]
\centering
\caption{Network Architectures - MLP1, Encoder, Decoder, MLP2.}
\label{tab:network_phi}
\begin{tabular}{|c|c|c|c|}
\hline
\textbf{Network} & \multicolumn{1}{|l|}{\textbf{Layer}} & \textbf{Width}   & \multicolumn{1}{l|}{\textbf{Activation}} \\ \hline
Inputs    & Cell    & 2      & \\ 
& 4 neighbors    & 8      & \\ \hline
& Input    & 8      & \\ 
& HL 1    & 8      & ReLU\\ 
MLP1   & HL 2    & 8      & ReLU\\ 
& Output  & 8      & tanh\\ \hline
& Input & 2,8      & \\ 
Concatenate & Output & 10      & \\ \hline
& Input    & 10      & \\ 
& HL 1    & 8      & ReLU\\ 
Encoder    & HL 2    & 8      & ReLU \\ 
& Output  & Latent Space (h) & ReLU \\ \hline
& Input    & Latent Space (h)      & \\ 
& HL 1    & 8      & ReLU\\ 
Decoder    & HL 2    & 8      & ReLU\\ 
& Output  & 2 & ReLU    \\ \hline
& Input    & Latent Space (h) & \\ 
& HL 1    & 8      & ReLU\\ 
MLP2  & HL 2    & 8      & ReLU\\ 
& Output  & 2      & tanh    \\ \hline
\end{tabular}
\end{table}

\subsubsection{Sparse Identification of Nonlinear Dynamical Systems (SINDy)}
\label{sec:SINDy}

The Sparse Identification of Nonlinear Dynamical Systems (SINDy) method is a data-driven approach used for discovering governing equations of dynamical systems from time-series data~\cite{sindy2016}. It aims to identify the underlying dynamics by expressing the system's behavior in terms of a sparse set of nonlinear functions. In our study, we use SINDy on the outputs generated by the learned CA models. We use a polynomial library of functions to identify parameters for the reaction part, alongside a Laplacian function for the diffusion part.

\subsubsection{Cellular Automata Simulation}
In our study, we use CA simulations using the learned CA models to predict dynamics starting from a randomized initial condition. In our experiments, we use the same initial condition as in the test observed data. The simulations are performed by recursively applying the update rule as mentioned in Section \ref{sec:CA}.

\begin{figure*}
    \centering
    \includegraphics[width=0.8\linewidth]{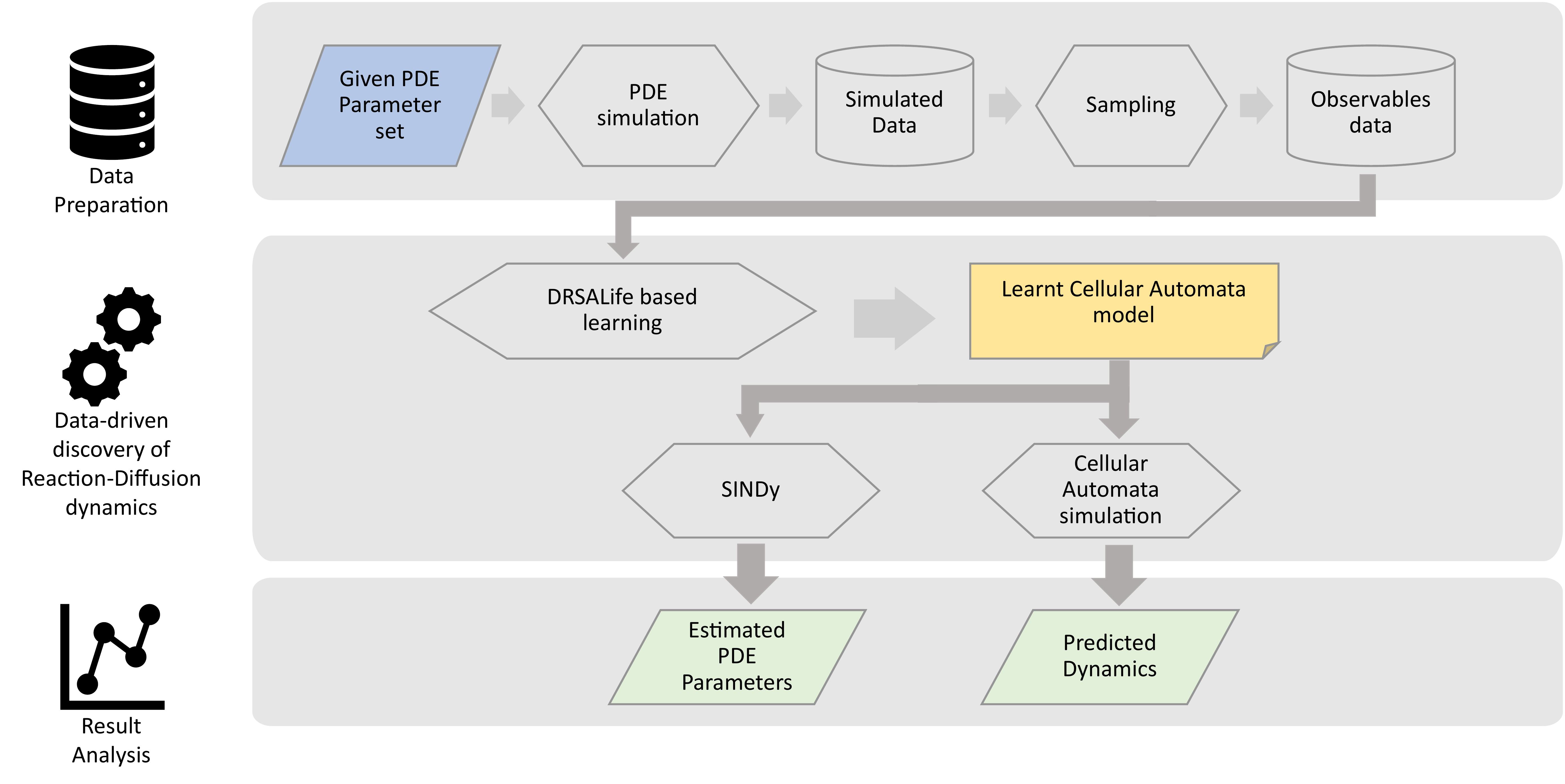}
    \caption{This figure depicts the methodology applied in our experiments to estimate PDE parameters with variations to observed data in terms of noise and sparsity. The methodology contains three main components: Data Preparation, Method application (Data-driven discovery of reaction diffusion (RD) dynamics), and Result analysis.}
    \label{fig:validation_workflow}
\end{figure*}

\subsection{Experimental Protocol}
\label{sec:methodology}
To evaluate our approach's ability to capture complex dynamics in terms of nonlinearity and emergence, we applied an experiment-based methodology. The first experiment aimed at evaluating the proposed method's ability to learn the dynamics of an RD system. The second experiment was designed to evaluate the method's robustness to noisy data. The third and fourth experiments were designed to evaluate the method's ability to learn from sparse data, with a varying nature of sparsity. In all experiments, we applied a methodology,  shown in Figure~\ref{fig:validation_workflow}, to estimate PDE parameters with respect to variations to training data in terms of noise and sparsity. This methodology contains three broad components: First, Data Preparation, in which we prepare and sample the simulated data to be used in experiments; Second, Method application, in which we apply the method described in Section \ref{sec:method} to the observed data; and third, Result analysis, in which we qualitatively and quantitatively measure the performance of the proposed method using evaluation metrics. 

\subsubsection{Dataset Description}
\label{sec:Dataset_decription}
The dataset used for training is simulated using the finite difference method \cite{ForsytheGeorgeE1960Fmfp,LeVeque2007}, with the FHN model given by PDE(s) defined by equations~\ref{eq:FHN_eq1} and \ref{eq:FHN_eq2} in the domain $E=[-1,1]^2$ 

\begin{equation}
    \label{eq:FHN_eq1}
    \frac{du}{dt} = u-u^{3} + k - v + a\Delta u
\end{equation}

\begin{equation}
    \label{eq:FHN_eq2}
    \tau\frac{dv}{dt} = u - v + b\Delta v
\end{equation}

\noindent where $u$ and $v$ represent the concentrations of activator and inhibitors, respectively, using the following parameters' values in the FHN model: $a = 2.8 \times 10^{-4}$, 
$b = 5 \times 10^{-3}$,
$\tau = 0.1$, and
$k = 0.005$. The parameter value set characterizes instabilities that cause labyrinth patterns \cite{Goldstein1996}.

For the training dataset, we used 10 simulations, each with a duration of 25 seconds and a time step of \( \Delta t = 0.001 \) seconds, resulting in 25{,}000 timesteps per simulation. This duration was sufficient for the system to reach equilibrium. Simulations were initialized with random initial conditions for \( u \) and \( v \). Observables were recorded every second, i.e., every 1{,}000 timesteps.

\subsubsection{Evaluation metrics}
\label{metrics}
To evaluate the performance of the method, we employed the following metrics to quantify differences between the observed and estimated state of the system:
\begin{itemize}
    \item \textbf{Structural Similarity Index measure (SSIM):} We used the SSIM metric as defined by \cite{SSIM2004} to calculate the structural similarity between the observed and estimated state. The SSIM metric is calculated as a combination of Luminance comparison, Contrast comparison, and Structure comparison \cite{SSIM2004}, as:

    \begin{equation}
        \text{SSIM}(x, y) = \frac{(2\mu_x\mu_y + C_1)(2\sigma_{xy} + C_2)}{(\mu_x^2 + \mu_y^2 + C_1)(\sigma_x^2 + \sigma_y^2 + C_2)}
    \end{equation}
    where,\\
    $x$ and $y$ are the two images being compared,\\
    $\mu_x$ and $\mu_y$ are the average luminance of the images,\\
    $\sigma_x^2$ and $\sigma_y^2$ are the variances (contrast),\\
    $\sigma_{xy}$ is the covariance between the two images,\\
    $C_1$ and $C_2$ are small constants to avoid division by zero.
    
    \item \textbf{Histogram based accuracy (HIST):} This metric is usually used to evaluate the similarity between two images by comparing the histograms of the images. In our experiments, we use this metric to assess the similarity between the observed and estimated state of the system. We use OpenCV \cite{itseez2015opencv} to calculate the metric, which calculates the difference between two image histograms ($H_1$ and $H_2$) as:
    \begin{equation}
        d(H_1,H_2) =  \frac{\sum_I (H_1(I) - \bar{H_1}) (H_2(I) - \bar{H_2})}{\sqrt{\sum_I(H_1(I) - \bar{H_1})^2 \sum_I(H_2(I) - \bar{H_2})^2}}
    \end{equation}
    where,
    \begin{equation}
        \bar{H_k} =  \frac{1}{N} \sum _J H_k(J)
    \end{equation}
    and is $N$ the total number of histogram bins.
    \item \textbf{Mean absolute error based accuracy (MAE\_Accuracy):} We used mean absolute error (MAE) to calculate this metric.
    \begin{equation}
        \begin{split}
            MAE = \frac{1}{N}\sum_{i=1}^{N}|X_i-Y_i| \\ MAE\_Accuracy = 1- MAE
        \end{split}
    \end{equation}
    where $X$ and $Y$ represent the observed and estimated state of the system with $N$ cells.
\end{itemize}

\subsubsection{Experimental setup}

\begin{enumerate}
    \item \textbf{Training Data with no noise and sparsity}
    
    The following process was adopted to perform the experiments. First, a parameter set for the RD system was selected, which involved defining specific values for key factors such as diffusion and reaction rates (see Section \ref{sec:Dataset_decription}). Next, simulated data for the RD system was generated based on a finite difference method. From this simulated dataset, observable data were then sampled, with sampling described in Section~\ref{sec:Dataset_decription}. Following this, the method described in Section~\ref{sec:method} was applied to learn the emergent dynamics from the observable data, to output a CA model. SINDy~\cite{sindy2016} was then utilized on both the observable data and the learned CA model (or ruleset), to estimate PDE parameters. A comparison was conducted between the initially selected parameter set and the estimated parameters obtained from the observable data and the learned CA model. Finally, a qualitative and quantitative comparison was performed between the steady/equilibrium state estimated using the learned CA model and the observable data, using metrics to assess their similarities (see Section \ref{metrics}).
    
    \item \textbf{Training Data with Gaussian noise}
    
    Next, we evaluated our method's robustness towards Gaussian noise. To perform this experiment, Gaussian noise was added at varying levels to the observable data. We introduced Additive White Gaussian Noise (AWGN) levels according to equation \ref{eq:snr} \cite{McClaningKevin2012WRDf} with Signal-to-Noise Ratio (SNR) values (in dB) of 100, 35, 30, 25, 10, and 1. Following this, the same process, as described in the first experiment, was followed. Here, one CA model per noise level was trained.

    \begin{equation}
    \label{eq:snr}
        \text{AWGN} = \mathcal{N}(0,\sigma^2)
    \end{equation}
    
    \[
    \sigma = \bar{S} \times 10^{(-SNR_{dB} / 10)}
    \]

    where, $\mathcal{N}$ represents the normal distribution with mean $0$ and standard deviation $\sigma$, $\bar{S}$ represents the average $u$ and $v$ concentrations and $SNR_{dB}$ is the signal-to-noise ratio value in decibels.

    \item \textbf{Training Data with Temporal sparsity} 
    
    Then, we evaluated our method's robustness towards temporal sparsity. To perform this experiment, temporal sparsity was added in varying levels to the observable data (we introduced random sparsity in terms of a fraction of the dataset comprising 80\%, 50\%, 40\%, 30\%, and 10\% of the observable data). Following this, the same process, as described in the first experiment, was followed. Here, one CA model per sparsity level was trained.

    \item \textbf{Training Data limited to the vicinity of the equilibrium state}
    
    Lastly, we evaluated our method's robustness towards limitations on data availability away from the equilibrium state. To perform this experiment, limitations on the observable data availability away from the equilibrium state were imposed in varying levels (we restricted training data in terms of maximum degree of average change in activator and inhibitor concentrations as $10^{-0.5}$ (less limitations on data availability), $10^{-1.0}$, $10^{-1.6}$, $10^{-1.8}$, $10^{-2.0}$, $10^{-2.2}$ (data restricted to very vicinity of equilibrium)). Following this, the same process, as described in the first experiment, was followed. Here, one CA model per limitation level was trained.
\end{enumerate}

For each experiment, we performed quantitative evaluations using metrics described in Section \ref{metrics} on 100 test simulations starting with random initial conditions. To train the models, we used a 16-core 64GB RAM machine. The proposed method was implemented using the Python-based TensorFlow-Keras library \cite{chollet2015keras}. Our code implementation of the proposed method, experiments, and results are provided online at \url{https://github.com/saumitrd92/DRSALife_RD}.

\begin{figure}[!t]
    \centering
    \begin{subfigure}[b]{\linewidth}
        \centering
        \fbox{\includegraphics[width=\linewidth]{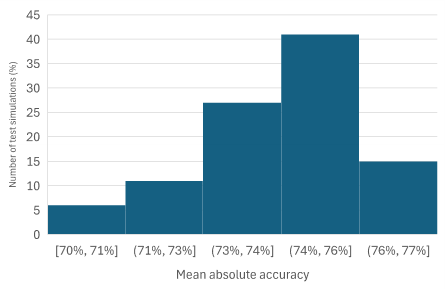}}
        \caption{}
        \label{fig:results_exp1_top}
    \end{subfigure}
    \begin{subfigure}[b]{\linewidth}
        \centering
        \fbox{\includegraphics[width=\linewidth]{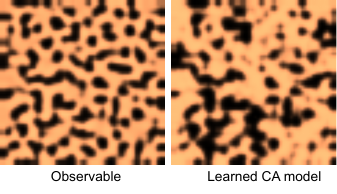}}
        \caption{}
        \label{fig:results_exp1_bottom}
    \end{subfigure}
    \caption{Figure \ref{fig:results_exp1_top} shows the mean absolute accuracy of 100 simulations (one of which is shown in the Figure \ref{fig:results_exp1_bottom}) using the learned CA model from the first experiment.}
    \label{fig:results_exp1}
\end{figure}

\begin{table*}[h!]
    \centering
    \caption{Tables present the parameters used to prepare the simulated data and the parameters estimated from test observables data (after sampling) and learned CA model for $\dot{u}$ and $\dot{v}$ as given in Equations \ref{eq:FHN_eq1} and \ref{eq:FHN_eq2}, respectively.}
    \label{tab:results_exp1}
    \begin{tabular}{cc}
        \begin{subtable}{0.52\linewidth}
            \caption{PDE parameters for $\dot{u}$}
            \label{tab:results_exp1_u}
            \centering
            \begin{tabular}{|l|r|r|r|}
                \hline
                \rowcolor{lightgray}
                \textbf{Parameters} & \textbf{$\dot{u}$} & \textbf{Observables} & \textbf{Learned CA}    \\ \hline
                \rowcolor{white}
                $\Delta u$  & $2.8\times 10^{-4}$   & $1.38\times 10^{-4}$  & $1.36\times 10^{-4}$      \\ \hline
                $\Delta v$  & 0                     & $-9.8\times 10^{-5}$  & $-7.95\times 10^{-5}$     \\ \hline
                $k$         & $5\times 10^{-3}$     & $-4.79\times 10^{-2}$ & $-5.6\times 10^{-2}$      \\ \hline
                $u$         & 1                     & 0.53                  & 0.50                      \\ \hline
                $v$         & -1                    & -0.44                 & -0.39                     \\ \hline
                $u^2$       & 0                     & $5.3\times 10^{-2}$   & $5.4\times 10^{-2}$       \\ \hline
                $v^2$       & 0                     & 0                     & 0                         \\ \hline
                $uv$        & 0                     & 0                     & 0                         \\ \hline
                $u^2v$      & 0                     & 0                     & 0                         \\ \hline
                $uv^2$      & 0                     & 0                     & 0                         \\ \hline
                $u^2v^2$    & 0                     & 0                     & 0                         \\ \hline
                $u^3$       & -1                    & -0.55                 & -0.52                     \\ \hline
                $v^3$       & 0                     & 0                     & 0                         \\ \hline
                $uv^3$      & 0                     & 0                     & 0                         \\ \hline
                $u^3v$      & 0                     & 0                     & 0                         \\ \hline
                $u^3v^2$    & 0                     & 0                     & 0                         \\ \hline
                $u^2v^3$    & 0                     & 0                     & 0                         \\ \hline
                $u^3v^3$    & 0                     & 0                     & 0                         \\ \hline
            \end{tabular}
        \end{subtable}
        \hspace{0.005\linewidth} % Space between subtables
        \begin{subtable}{0.40\linewidth}
            \centering
            \caption{PDE parameters for $\dot{v}$}
            \label{tab:results_exp1_v}
            \begin{tabular}{|l|r|r|r|}
                \hline
                \rowcolor{lightgray}
                \textbf{Parameters} & \textbf{$\dot{v}$} & \textbf{Observables} & \textbf{Learned CA}    \\ \hline
                \rowcolor{white}
                $\Delta u$  & 0                 & $4.2\times 10^{-5}$   & $3.9\times 10^{-5}$           \\ \hline
                $\Delta v$  & $5\times 10^{-2}$ & $1.03\times 10^{-5}$  & $6.1\times 10^{-5}$           \\ \hline
                $k$         & 0                 & 0                     & $-2.3\times 10^{-3}$          \\ \hline
                $u$         & 10                & 0.16                  & 0.16                          \\ \hline
                $v$         & -10               & -0.28                 & -0.20                         \\ \hline
                $u^2$       & 0                 & $2.02\times 10^{-2}$  & $3.3\times 10^{-2}$           \\ \hline
                $v^2$       & 0                 & -0.26                 & -0.37                         \\ \hline
                $uv$        & 0                 & 0                     & 0                             \\ \hline
                $u^2v$      & 0                 & 0                     & 0                             \\ \hline
                $uv^2$      & 0                 & 0                     & 0                             \\ \hline
                $u^2v^2$    & 0                 & 0                     & 0                             \\ \hline
                $u^3$       & 0                 & 0                     & $-2.2\times 10^{-2}$          \\ \hline
                $v^3$       & 0                 & 0                     & 0                             \\ \hline
                $uv^3$      & 0                 & 0                     & 0                             \\ \hline
                $u^3v$      & 0                 & 0                     & 0                             \\ \hline
                $u^3v^2$    & 0                 & 0                     & 0                             \\ \hline
                $u^2v^3$    & 0                 & 0                     & 0                             \\ \hline
                $u^3v^3$    & 0                 & 0                     & 0                             \\ \hline
            \end{tabular}   
        \end{subtable}
    \end{tabular}
\end{table*}

\section{Results}
\label{sec:results}

In this section, we present outcomes from our experiments. We compile and present statistics, using a histogram and box-whisker plots, for the three evaluation metrics. We specifically present findings on the prediction accuracy of the learned models, trained using observed data with varying configurations on noise and sparsity. We also tabulate the identified PDE parameters using the learned models.

\subsection{Parameter estimation}

We quantitatively and qualitatively compare the observed (using the PDE model) and predicted (using the learned CA model) system state at $t=8$ seconds. In Figure \ref{fig:results_exp1_top}, the histogram presents the mean absolute accuracy of \textbf{74\%}, in 100 test simulations (each estimated to $t=8$ seconds with random initial conditions). In Figure \ref{fig:results_exp1_bottom}, we depict the observed vs predicted system state at $t=8$ seconds, to qualitatively show the similarities in the pattern textures.  Moreover, Table~\ref{tab:results_exp1} details the PDE parameters estimated using the learned CA model and the observable data. Here, we can observe the similarities in the estimated parameters from the observable data and the learned CA model. For instance, with regards to $\dot{u}$, the estimated diffusivity coefficients (parameters for $\Delta u$) and reaction coefficients (parameters for $u^3$, $u^2$, $u$, and $v$) are quite comparable for Observables and learned CA model given in the third and fourth columns of Table~\ref{tab:results_exp1_u}. However, these estimated parameters (third and fourth columns) are quite different from the original parameters used in Section~\ref{sec:Dataset_decription} (also given in the second column of this table), as they are estimated from dynamics sampled at a low rate (sampling rate mentioned in Section \ref{sec:Dataset_decription}).

\subsection{Robustness to Gaussian noise}

Figure \ref{fig:results_exp2} details the prediction accuracy of the trained CA model across varying levels of Gaussian noise in observed data. In this figure, we use box-whisker plots to present model accuracy (using three metrics) across varying configurations of Gaussian noise in observed data used to train the CA models. We can observe that the accuracy of the learned CA models is high for noise levels SNR$\geq$25, and decreases with noise levels SNR$\leq$10.

\subsection{Robustness to Temporal sparsity noise}

\begin{figure}[!t]
    \centering
    \includegraphics[width=\linewidth]{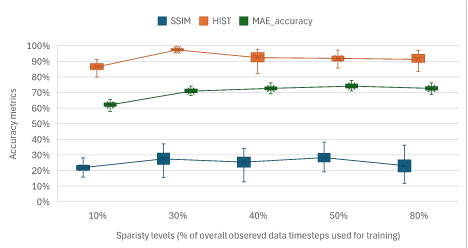}
    \caption{Figure shows the accuracy of the learned CA models subjected to varying degrees of temporal sparsity.} 
    \label{fig:results_exp3}
\end{figure}

Figure \ref{fig:results_exp3} details the prediction accuracy of the trained CA model across varying temporal sparsity in observed data. In this figure, we use box-whisker plots to present model accuracy (using three metrics) across varying configurations of temporal sparsity in observed data used to train the CA models. The accuracy of the learned CA models is high for sparsity $\geq$30\% of observable data, and lowers with sparsity $\leq$10\% of observable data.

\subsection{Robustness to training data limited to equilibrium}

\begin{figure}[!t]
    \centering
    \includegraphics[width=\linewidth]{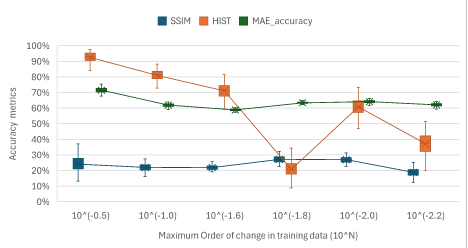}
    \caption{Figure shows the accuracy of the learned CA models subjected to varying degrees of data availability towards the equilibrium state.}
    \label{fig:results_exp4}
\end{figure}

Figure \ref{fig:results_exp4} details the prediction accuracy of the trained CA model across varying observability, far from the equilibrium state. In this figure, we use box-whisker plots to present model accuracy (using three metrics) across varying configurations of observability in observed data used to train the CA models. The accuracy of the learned CA models lower drastically when observed data points are limited towards near-equilibrium states.

\begin{figure}[!t]
    \centering
    \includegraphics[width=\linewidth]{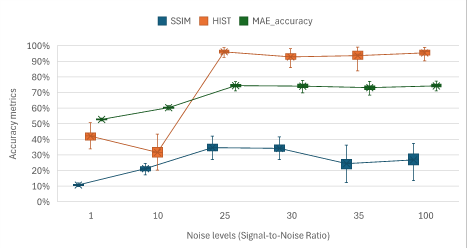}
    \caption{The accuracy of the learned CA models subjected to varying degrees of Gaussian noise.} 
    \label{fig:results_exp2}
\end{figure}

\section{Discussions}
\label{sec:Discussions}

\textcite{Noordijk2024} provide a comprehensive overview of challenges in using Machine learning (ML) and Mechanistic modeling separately in systems biology. For instance, according to \textcite{Noordijk2024}, the mechanistic models struggle with the high dimensionality of the biological processes coming from a large number of parameters and involved complex interactions. On the other hand, ML models struggle with overfitting, especially with noisy and limited data, and the lack of interpretability regarding the underlying biological mechanisms. Reviews like \cite{Noordijk2024} highlight the need for innovative methodologies that can effectively combine mechanistic modeling with data-driven techniques to advance our understanding of system dynamics. With the approach proposed in this paper, we address such challenges pertaining to the modeling of spatiotemporal emergent behaviors by combining Soft ALife ruleset learning and system identification.

\subsection{Soft ALife rule-set learning}
There is a significant challenge associated with Soft ALife techniques, such as CA and ABM, in identifying or defining a ruleset that accurately reflects the dynamics of a system. In this paper, we present the applicability of the DRSALife model for RD emergent dynamics, a model that has shown promising results pertaining to Koopman-based learning of SoftALife rulesets for Elementary CA Rule 30, Game of Life, and Vicsek Flocking \cite{Dwivedi2025}. The experimental findings reported in this paper indicate that the proposed DRSALife-based method has the ability to effectively capture universal nonlinear behaviors in an RD system, while additionally providing valuable insights into system dynamics. In contrast to \cite{Dwivedi2025}, where the Koopman-based linearization is used, in this paper, we investigate the use of ANN-based feature transformation to learn the SoftALife rulesets. Through our experiments on an RD system, we find that it is possible to reliably map the low-level system dynamics into a high-dimensional hyperspace. The findings imply that the DRSALife-based method is able to learn a representative CA ruleset that models the spatiotemporal dynamics of the emergent system. The learned CA model, as shown in Figure~\ref{fig:results_exp1}, predicts the emergent dynamics with an average of 73-75\% accuracy on test simulations.

Moreover, as shown in Figures~\ref{fig:results_exp2} and \ref{fig:results_exp3}, the learned CA models prove reasonably robust to Gaussian noise and temporal sparsity, respectively. The reason behind such robustness to noise and temporal sparsity can be argued with reference to the architecture of the DRSALife model, as it uses several instances/data-points of low-scale dynamics in training data to learn aggregate representative dynamics.

\subsection{System Identification}

Regarding estimation of PDE parameters, SINDy \cite{sindy2016} is used on the observables and learned CA models. When we compare the PDE parameters estimated from observed data (see observables columns in Table \ref{tab:results_exp1}) and learned CA models (see learned CA columns in Table \ref{tab:results_exp1}), we find that the structure of the PDE is preserved. Moreover, we find that the estimated parameters from learned CA models are quite similar to the ones estimated from observed data. However, the challenge of observability remains, since the estimated PDE parameters using the learned CA models (see learned CA columns in Table \ref{tab:results_exp1}) and the observed data (see observables columns in Table \ref{tab:results_exp1}) differ from the original PDE parameters used in Section \ref{sec:Dataset_decription} (see $\dot{u}$ and $\dot{v}$ columns in Table \ref{tab:results_exp1}). This is due to a low sampling rate used to observe the data.

\subsection{Data availability and validity}

Our method relies on data-driven techniques, which necessitate access to data that accurately reflects the system's behavior. For example, if the data is lacking or only partially observed, particularly regarding the long-term behavior of the system, our approach will struggle to learn those behaviors. This is quite evident in our experiment, where the training data was limited to the equilibrium state of the system (Figure \ref{fig:results_exp4}). Overall, our findings indicate the robustness of our proposed method to noise, temporal sparsity, and limited measurement-scope data. Additionally, in our experiments, we limited the sampling of the observed data to induce characteristics of a real dataset.

Another aspect of data availability is the use of simulated data. In our experiments, simulations to produce the data needed for model training were utilized. This led to complete control in the experimental setup over the data availability for both modeling and validation purposes.

\subsection{Implications to practice and research}

In the upcoming sections, we will answer our research question by summarizing the implications of our findings for both practical applications and further research. We reiterate our research question here and answer it in terms of two broad implications, i.e. pertaining to Data-driven modeling and discovery of emergent RD dynamics and Inferences drawn from varying noise, sparsity configurations in data.\\

\noindent\textbf{RQ: What are the implications of using the DRSALife model for learning the underlying dynamics of a reaction-diffusion system and identifying the parameters of a representative partial differential equation?}

\subsubsection{Data-driven modeling and discovery of emergent RD dynamics}

Through our experiments, the DRSALife-based method has demonstrated its ability to learn emergent behavior from observed data for a reaction diffusion system. We show that the learned CA models show reasonable predictability even with high levels of sparsity and noise in the training data. As this proposed method demonstrates a robust way to model CA rulesets using observed RD data, without any assumptions/knowledge on system dynamics, it can be highly useful to practitioners and researchers like \textcite{WANG2024938,Hou2025} working with RD systems. For instance, in \cite{WANG2024938}, the authors propose an innovative approach to regulate the self-assembly of supramolecular hydrogels through the RD process, providing valuable insights for the development of advanced materials that exhibit lifelike characteristics and functionalities. Our proposed approach, presented in this paper, can be highly useful in such research wherein the data gathered from experiments of the RD process can be used to create representative Soft ALife models like ABM or CA to predict and control the self-assembly of supramolecular hydrogels with varying spatiotemporal parameters like concentrations of urease and shapes of the diffusion fronts. Similar applicability of our proposed method is possible in research like \cite{Hou2025}, wherein the authors examine the dynamics of vegetation patterns using the framework of optimal control theory, emphasizing its significance in ecological management and restoration efforts. Through our proposed method, data-driven Soft ALife models can be developed for vegetation dynamics to enhance the understanding and management of vegetation changes. 

The proposed method has also demonstrated its ability to work with methods like SINDy, to be used to identify the dynamics in the form of estimation of a representative PDE. This ability works well with regard to model transparency and helps to effectively understand the nature of system dynamics, which is highly useful to researchers and practitioners like \textcite{ABBAS2025386} working with mechanistic models, wherein the authors develop various PDE models for the dynamics of vegetation biomass in relation to autotoxicity. Using our proposed method, such PDE models can be based on observed data, thereby incorporating the required number of spatiotemporal variables and their interactions into the modeling of vegetation dynamics. 

Additionally, the proposed method works with a variety of data-driven methods such as ANN, kernel-based methods, Koopman-based methods \cite{Dwivedi2025}, and more. In other words, one can use and experiment with different kinds of data-driven methods to learn behavior at low scales while being able to choose the nature of the Soft ALife method (CA, ABM, etc.) at the high scale. This will potentially give high flexibility and applicability to the DRSALife-based method across various application domains/disciplines working with RD systems.

\subsubsection{Inferences from data configurations}
In the realm of machine learning, noise and sparsity in training data are widely recognized as significant challenges that can affect the quality of learned models. Our experiments reveal noteworthy implications concerning training data. Firstly, while the DRSALife-based method demonstrates considerable robustness to noise and sparsity, the accuracy of the learned models diminishes beyond certain thresholds of these factors. Secondly, the learned CA models exhibit a strong sensitivity to training data that is restricted to equilibrium states.

\section{Conclusions}
Emergent spatiotemporal dynamical systems, such as RD systems, are widely studied in several subject areas like neuroscience, ecology, chemistry, epidemiology, and more. However, there is limited research into learning Soft Artificial Life models, like Agent-based and Cellular Automata models, from observed data for RD systems. Additionally, data-driven discovery without guidance from prior knowledge of underlying physics still remains a challenge. 

In this paper, we present implications of using the DRSALife conceptual model for learning the dynamics of an RD system from observed data. We present findings from our experiments, investigating the predictability and robustness of the learned CA models to varying configurations of Gaussian noise and temporal sparsity in observed data. Moreover, we present implications of using SINDy~\cite{sindy2016} along with the learned CA models to estimate PDE parameters, thereby making the learned CA models transparent, wherein the estimated PDE parameters provide insight into the structure of learned emergent dynamics. Findings show that the learned CA models are accurately able to predict the emergent dynamics and are quite robust to noise and sparsity in observed data. Additionally, findings show that the estimated PDE parameters from the learned CA models are quite similar in structure and values to the ones estimated directly from observed data. The findings presented in this paper may offer valuable insights for researchers and practitioners. Firstly, one can use the proposed method's ability to learn Soft ALife rule-sets from observed RD data while also taking account of its robustness to noise and sparsity in data. Moreover, one can use the proposed method along with methods like SINDy~\cite{sindy2016} to identify system dynamics. Also, one can exploit the inherent flexibility of the DRSALife model's architecture to choose the kind of data-driven method to learn from observed data at low scale, as well as the nature of Soft ALife method (ABM, CA, or some else) at the high scale. This flexibility points to additional avenues for future research. In our research presented in this paper, we work with CA models learned using ANN-based learning. One can also investigate the applicability of other data-driven methods like kernel-based learning to model emergent behavior alongside other Soft ALife methods like Agent-based models, artificial chemistry and digital phenotyping.

\section*{Acknowledgments/Funding}
This research did not receive any specific grant from funding agencies in the public, commercial, or not-for-profit sectors. However, it was supported by institutional resources provided by the Norwegian University of Science and Technology (NTNU).\footnote{Norwegian University of Science and Technology (NTNU) --  \url{https://www.ntnu.edu/} (As of May 2025).}

\section*{Competing Interests/Conflict of interest}
The authors declare that they have no known competing financial interests or personal relationships that could have appeared to influence the work reported in this paper.

\section*{Ethical Approval}
This study did not involve any human participants, animals, or data derived from them.

\section*{Author contribution}
\textbf{Saumitra Dwivedi:} Conceptualization, Data Curation, Investigation, Methodology, Software, Formal analysis, Visualization, Validation, Writing - Original Draft. \textbf{Ricardo da Silva Torres:} Conceptualization, Methodology, Validation, Writing - Review \& Editing, Supervision. \textbf{Ibrahim A. Hameed:} Conceptualization, Methodology, Validation, Writing - Review \& Editing, Supervision. \textbf{Gunnar Tufte:} Conceptualization, Methodology, Validation, Writing - Review \& Editing, Supervision. \textbf{Anniken Susanne T. Karlsen:} Conceptualization, Methodology, Validation, Writing - Review \& Editing, Supervision, Project administration.

\section*{Code availability and Data availability}
Our code implementation of the proposed method, alongside training data and results of experiments, is provided online at \url{https://github.com/saumitrd92/DRSALife_RD}.

\printbibliography[title={References}]

\clearpage

\begin{IEEEbiography}[{\includegraphics[width=1in,height=1.25in,clip,keepaspectratio]{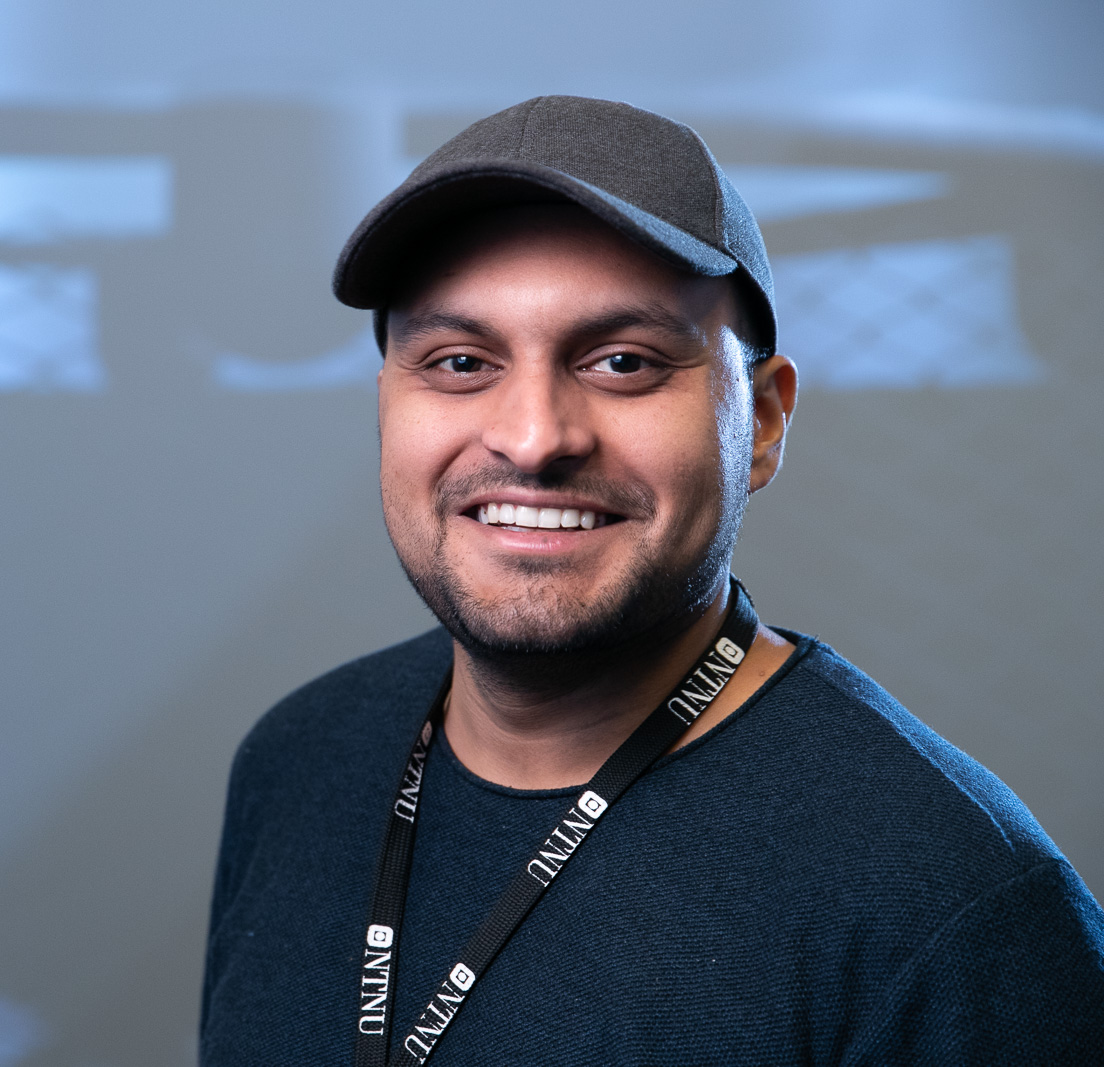}}]{Saumitra Dwivedi} is currently a PhD candidate at the Department of ICT and Natural Sciences, Faculty of Information Technology and Electrical Engineering, Norwegian University of Science and Technology (NTNU), Ålesund, Norway. He received a M.Sc. in Simulation and Visualization from Norwegian University of Science and Technology (NTNU) and a B.Tech in Petroleum Engineering from Indian Institute of Technology (IIT-ISM) Dhanbad, India.
\end{IEEEbiography}

\begin{IEEEbiography}[{\includegraphics[width=1in,height=1.25in,clip,keepaspectratio]{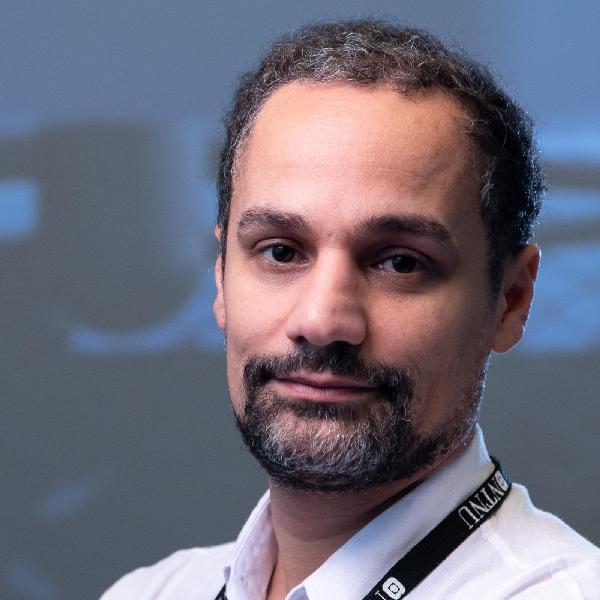}}]{Dr. Ricardo da Silva Torres} (IEEE Member) received his B.Sc. degree in Computer Engineering from the University of Campinas, Brazil, in 2000, and his Ph.D. degree in Computer Science from the same university in 2004. He is currently a Professor of Data Science and Artificial Intelligence at Wageningen University and Research. From 2019 to 2024, he held a position as a Professor in Visual Computing at the Norwegian University of Science and Technology (NTNU). Prior to that, he was a Professor at the University of Campinas, Brazil, from 2005 to 2019. Dr. Torres has been developing multidisciplinary e-science research projects that focus on multimedia analysis, multimedia retrieval, machine learning, databases, and information visualization. He is the author or co-author of over 200 articles published in refereed journals and conferences and serves as a program committee member for several international and national conferences. He is currently serving as an Associate Editor for Pattern Recognition and Pattern Recognition Letters.
\end{IEEEbiography}

\begin{IEEEbiography}[{\includegraphics[width=1in,height=1.25in,clip,keepaspectratio]{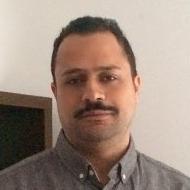}}]{Dr. Ibrahim A. Hameed} is a full professor and head of the Robotics Research Group at the Norwegian University of Life Sciences (NMBU) in Norway. He holds a BSc and MSc in Industrial Electronics and Control Engineering from Menoufia University, Egypt, a PhD in Industrial Systems and Information Engineering from Korea University, South Korea, and a PhD in Mechanical Engineering from Aarhus University, Denmark. Hameed is a Senior Member of IEEE and the former chair of the IEEE Computational Intelligence Society (CIS) Norway Section. His research interests include artificial intelligence, machine learning, optimization, control systems, and robotics.
\end{IEEEbiography}

\begin{IEEEbiography}[{\includegraphics[width=1in,height=1.25in,clip,keepaspectratio]{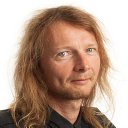}}]{Dr. Gunnar Tufte} is Professor at the Department of Computer Science, Faculty of Information Technology and Electrical Engineering at the Norwegian University of Science and Technology (NTNU). Dr. Tufte holds the positions of Deputy Head of Department (Research), and Head of the PhD Program in the Department of Computer Science (IDI). He is part of the Computer Architecture Lab (CAL) and the Computational Magnetic Metamaterials (COMET) research groups at NTNU. His research interests include Unconventional Machines: Architecture, Design and Computation, Artificial Life, Evolution-in-Materio, Complex Systems, and Nanosystems.
\end{IEEEbiography}

\begin{IEEEbiography}[{\includegraphics[width=1in,height=1.25in,clip,keepaspectratio]{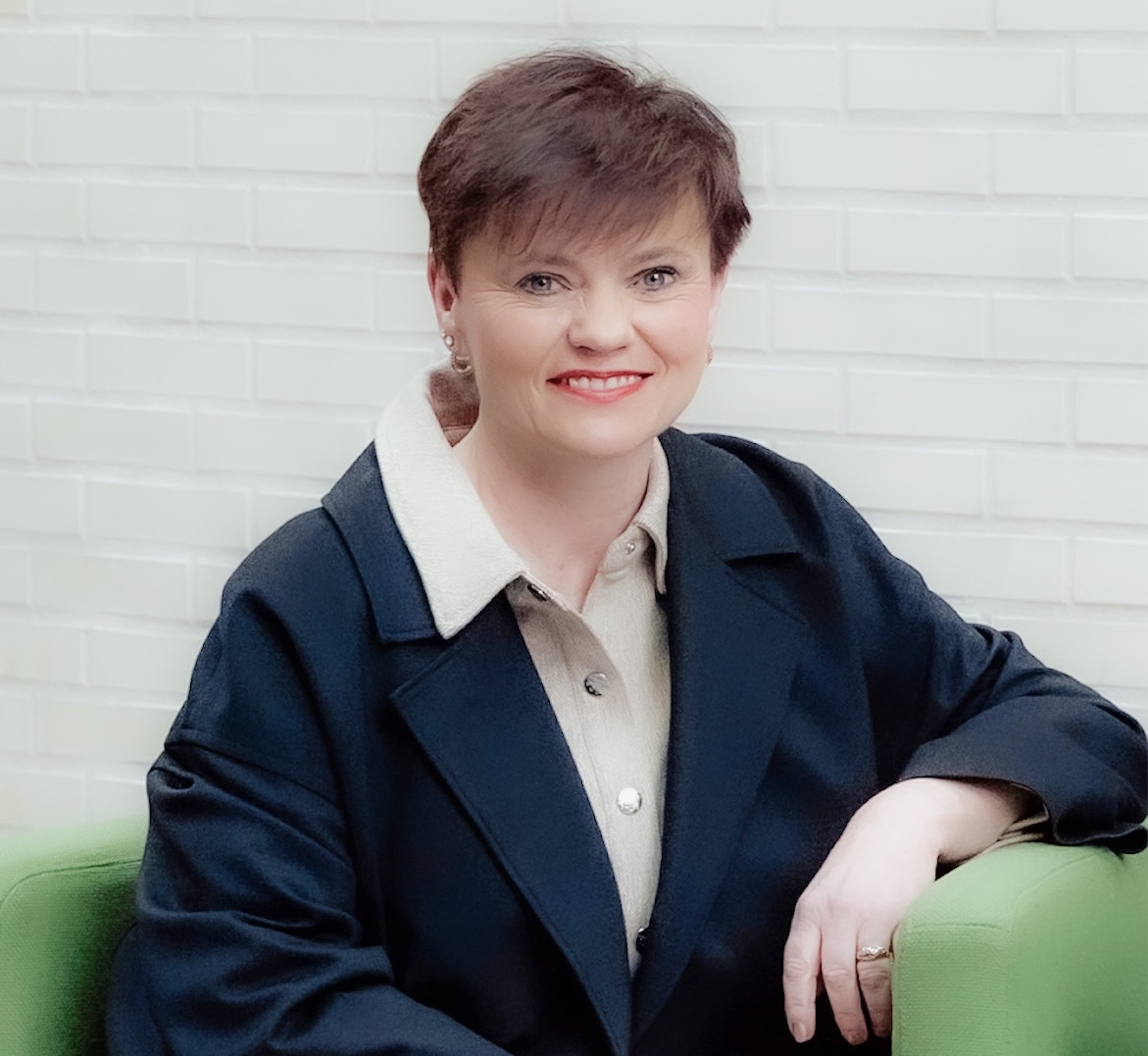}}]{Dr. Anniken Susanne T. Karlsen} researches and lectures in Computer \& Information Sciences (CIS) at the Norwegian University of Science and Technology (NTNU). She earned her PhD in Information Science from the University of Bergen (UiB) and holds a Master’s degree in Information Technology from Aalborg University (AAU) in Denmark. She is also a Siviløkonom from the Norwegian School of Economics (NHH), and a Computer Engineer from the former Møre and Romsdal University College of Engineering (MRIH). Dr. Karlsen's research and teaching interests in the field of Computer \& Information Sciences (CIS) span a wide range of topics, from fundamental theory to practical technology applications. Her work is centrally focused on the design and development of advanced systems through the integration of concepts from diverse domains. Her interests include modern software engineering, socio-technical systems thinking, systems engineering, artificial intelligence, computer simulation and complexity.
\end{IEEEbiography}

\EOD

\end{document}